\documentclass{article}
\usepackage{amssymb}
\usepackage{amsfonts}
\usepackage{amsmath}

\newtheorem{theorem}{Theorem}

\newtheorem{lemma}[theorem]{Lemma}

\newenvironment{proof}[1][Proof]{\noindent\textbf{#1.} }{\ \rule{0.5em}{0.5em}}
\input{tcilatex}

\begin{document}

\title{A Note on the PAC Bayesian Theorem}
\author{Andreas Maurer \\
Adalbertstr. 55\\
D-80799 M\"{u}nchen\\
\textit{andreasmaurer@compuserve.com}}
\maketitle

\begin{abstract}
We prove general exponential moment inequalities for averages of
[0,1]-valued iid random variables and use them to tighten the PAC Bayesian
Theorem. The logarithmic dependence on the sample count in the enumerator of
the PAC Bayesian bound is halved.
\end{abstract}

\section{Introduction}

The relative entropy or Kullback Leibler divergence of a Bernoulli variable
with bias $p$ to a Bernoulli variable with bias $q$ is given by%
\begin{equation*}
KL\left( p,q\right) =p\ln \frac{p}{q}+\left( 1-p\right) \ln \frac{1-p}{1-q}.
\end{equation*}%
Suppose that $\mathbf{X}=\left( X_{1},...,X_{n}\right) $ is a vector of iid
random variables, $0\leq X_{i}\leq 1$, $E\left[ X_{i}\right] =\mu $ and let $%
M\left( \mathbf{X}\right) =\left( 1/n\right) \sum X_{i}$ be the arithmetic
mean. We will derive the following inequality, valid for $n\geq 8$:%
\begin{equation}
E\left[ e^{nKL\left( M\left( \mathbf{X}\right) ,\mu \right) }\right] \leq 2%
\sqrt{n}  \label{simple upper bound}
\end{equation}%
We also show that the square root on the right side gives the optimal order
in $n$ because for Bernoulli ($\left\{ 0,1\right\} $-valued) variables $%
X_{i} $ we have the additional inequality, valid for $n\geq 2$,%
\begin{equation}
\sqrt{n}\leq E\left[ e^{nKL\left( M\left( \mathbf{X}\right) ,\mu \right) }%
\right] \text{.}  \label{simple lower bound}
\end{equation}%
We will also see that for Bernoulli variables the right side is independent
of $\mu $, so that the expectation $E\left[ e^{nKL\left( M\left( \mathbf{X}%
\right) ,\mu \right) }\right] $ is the same for all Bernoulli variables and
depends only on $n$. \bigskip

It is likely that the inequalities (\ref{simple upper bound}) and (\ref%
{simple lower bound}) are known. The upper bound (\ref{simple upper bound})\
can be applied to improve on the PAC-Bayesian Theorem (see e.g. \cite%
{McAllester 1999},\cite{McAllester 2003},\cite{Seger 2002}) in learning
theory: Suppose one has a set of data $\mathcal{Z}$ with probability measure 
$D$ and a set $\mathcal{H}$ of hypotheses $h:\mathcal{Z}\rightarrow \left[
0,1\right] $ (this already includes the usual loss-function). Suppose
further that there is a ('prior') probability measure $P$ on $\mathcal{H}$
(assume $\mathcal{Z}$ and $\mathcal{H}$ to be finite to avoid questions of
measurability). Then for any $\delta >0$, with probability greater than $%
1-\delta $ a sample $\mathbf{S=}\left( Z_{1},...,Z_{n}\right) \in \mathcal{Z}%
^{n}$ is drawn from $D^{n}$ such that for all ('posterior') probability
measures $Q$ on $\mathcal{H}$ we have for $n\geq 2$%
\begin{equation}
KL\left( E_{h\thicksim Q}\left[ M\left( h\left( \mathbf{S}\right) \right) %
\right] ,E_{h\thicksim Q}\left[ E_{z\thicksim D}\left[ h\left( z\right) %
\right] \right] \right) \leq \frac{KL\left( Q,P\right) +\ln \frac{1}{\delta }%
+\ln \left( 2n\right) }{n-1}.  \label{PAC Bayes bound}
\end{equation}%
Here $h\left( \mathbf{S}\right) $ refers to the vector $h\left( \mathbf{S}%
\right) =\left( h\left( Z_{1}\right) ,...,h\left( Z_{n}\right) \right) $, so
that $M\left( h\left( \mathbf{S}\right) \right) $ is the empirical loss of
the hypothesis $h$. The expression $KL\left( Q,P\right) $ refers to the
relative entropy of the probability measures $Q$ and $P$ (see \cite%
{CoverThomas}). The importance to learning theory comes from the fact that $%
Q $ may depend on $\mathbf{S}$. Note that (\ref{PAC Bayes bound}) implies 
\begin{equation*}
E_{h\thicksim Q}\left[ E_{z\thicksim D}\left[ h\left( z\right) \right] %
\right] \leq \sup \left\{ \epsilon :KL\left( E_{h\thicksim Q}\left[ M\left(
h\left( \mathbf{S}\right) \right) \right] ,\epsilon \right) \leq \frac{%
KL\left( Q,P\right) +\ln \frac{2n}{\delta }}{n-1}\right\} ,
\end{equation*}%
which can drive a learning algorithm to select a posterior $Q$ by minimizing
the sample-dependent right side . Among other applications (\cite%
{LangfordSeger 2002}, \cite{Seger 2002})\ the PAC Bayesian bound has been
applied to prove generalisation error bounds for large margin classifiers
such as support vector machines (\cite{LangfordSTaylor 2002}, \cite%
{McAllester 2003}).

The right side of (\ref{PAC Bayes bound}) has, with an overall factor of $%
1/\left( n-1\right) $, three terms: There is the relative entropy $KL\left(
Q,P\right) $, which can be interpreted as the information gain in
specializing from $P$ to $Q$, an information normally extracted from the
sample $\mathbf{S}$. The term $\ln \left( 1/\delta \right) $ expresses the
usual dependence on the confidence parameter $\delta $, but the remaining $%
\ln \left( 2n\right) $ is difficult to understand: Why do we need it, can't
it be altogether eliminated or at least reduced?

We do not know the answer to the first two questions, but using (\ref{simple
upper bound}) we can essentially cut the term in half, replacing $\ln \left(
2n\right) $ by $\ln \left( 2\sqrt{n}\right) $ for $n\geq 8$ and reduce the
overall factor to $1/n$. Our substitute for (\ref{PAC Bayes bound}) then
reads 
\begin{equation}
KL\left( E_{h\thicksim Q}\left[ M\left( \mathbf{S}\right) \right]
,E_{h\thicksim Q}\left[ E_{z\thicksim D}\left[ h\left( z\right) \right] %
\right] \right) \leq \frac{KL\left( Q,P\right) +\ln \frac{1}{\delta }+\ln
\left( 2\sqrt{n}\right) }{n}.  \label{PAC Bayes bound Tight}
\end{equation}

Our improvement is not spectacular, but significant when viewed in terms of
the confidence parameter $\delta $. It gives a slightly smaller
generalisation error bound (factor $\left( n-1\right) /n$) than (\ref{PAC
Bayes bound}) with a failure probability $\delta $ decreased by the factor $%
1/\sqrt{n}$. For example, if $n=10000$ and (\ref{PAC Bayes bound})\ gives a
fixed bound with a failure probability of $1/100$, our result will give the
same bound with failure probability less than $1/10000$.\bigskip 

It is possible to prove bounds similar to the above (see \cite{Catoni}\ and 
\cite{Audibert}), where the $\ln \left( n\right) $ dependence is replaced by 
$\ln \left( \ln \left( n\right) \right) $ or eliminated alltogether, at the
expense of multiplying $KL\left( Q,P\right) $ with a constant larger than
unity. The relative entropy $KL\left( Q,P\right) $ however is dependent on
the posterior $Q$ and thus implicitely on the sample and the sample-size $n$%
. In all cases where $KL\left( Q,P\right) $ grows faster than
logarithmically in $n$ (the generic case in machine learning) these bounds
will therefore be weaker than (\ref{PAC Bayes bound Tight}) above. 

We will prove the principal bounds (\ref{simple upper bound}) and (\ref%
{simple lower bound}) in section \ref{Section Main Bounds}. We will then
apply them to the PAC Bayesian Theorem in section \ref{Section PAC Bayes}%
.\bigskip

\section{Main Inequalities\label{Section Main Bounds}}

Throughout this note $X_{1},...,X_{n}$ are assumed to be IID random
variables with values in $\left[ 0,1\right] $ and expectation $E\left[ X_{i}%
\right] =\mu $. We use $\mathbf{X}$ to denote the corresponding random
vector $\mathbf{X}=\left( X_{1},...,X_{n}\right) $ with values in $\left[ 0,1%
\right] ^{n}$ and $M\left( \mathbf{X}\right) $ to denote its arithmetic mean%
\begin{equation*}
M\left( \mathbf{X}\right) =\frac{1}{n}\sum_{i=1}^{n}X_{i}.
\end{equation*}%
For any $\left[ 0,1\right] $-valued random variables $X$ use $X^{\prime }$
to denote the unique Bernoulli ($\left\{ 0,1\right\} $-valued) random
variable with $\Pr \left\{ X^{\prime }=1\right\} =E\left[ X^{\prime }\right]
=E\left[ X\right] $. Evidently $X^{\prime \prime }=X^{\prime }$, $\forall X$%
. For $\mathbf{X}=\left( X_{1},...,X_{n}\right) $ we denote $\mathbf{X}%
^{\prime }=\left( X_{1}^{\prime },...,X_{n}^{\prime }\right) $.

We restate our principal bounds in a slightly more general way.

\begin{theorem}
\label{Main Theorem}For all $n\geq 2$%
\begin{equation}
E\left[ e^{nKL\left( M\left( \mathbf{X}\right) ,\mu \right) }\right] \leq E%
\left[ e^{nKL\left( M\left( \mathbf{X}^{\prime }\right) ,\mu \right) }\right]
\leq e^{\frac{1}{12n}}\sqrt{\frac{\pi n}{2}}+2.  \label{upper bound}
\end{equation}%
If the $X_{i}$ are nontrivial Bernoulli variables (i.e. if $\mu \in \left(
0,1\right) $) then there is a sequence $c_{n}$ such that $1\leq
c_{n}\rightarrow \pi $ as $n\rightarrow \infty $ and%
\begin{equation}
e^{-\frac{1}{6}}\sqrt{\frac{n}{2\pi }}c_{n}+2\leq E\left[ e^{nKL\left(
M\left( \mathbf{X}\right) ,\mu \right) }\right] .  \label{lower bound}
\end{equation}%
In this case the expectation on the right is independent of $\mu $.\bigskip
\end{theorem}

The right side of (\ref{upper bound}) is bounded above by $2\sqrt{n}$ for $%
n\geq 8$ and the left side of (\ref{lower bound}) is bounded below by $\sqrt{%
n}$ for $n\geq 2$, thus giving the simpler bounds (\ref{simple upper bound})
and (\ref{simple lower bound}) of the introduction.\bigskip

To prove Theorem \ref{Main Theorem} we need some auxilliary results. The
first is Stirling's Formula:

\begin{theorem}
For $n\in \mathbb{N}$ 
\begin{equation}
n!=\sqrt{2\pi n}\left( \frac{n}{e}\right) ^{n}e^{\frac{g\left( n\right) }{12n%
}}  \label{Stirling}
\end{equation}%
with $0<g\left( n\right) <1$.
\end{theorem}

For a proof see e.g. \cite{Bauer 2002}. We will use this Theorem in form of
the following inequalities%
\begin{equation}
\sqrt{2\pi n}\left( \frac{n}{e}\right) ^{n}<n!<\sqrt{2\pi n}\left( \frac{n}{e%
}\right) ^{n}e^{\frac{1}{12n}}.  \label{Stirling II}
\end{equation}%
\bigskip

The following simple lemma shows that the expectation of a convex function
of iid variables can always be bounded by the expectation of the
corresponding Bernoulli variables.

\begin{lemma}
\label{Convexity lemma}Suppose that $f:\left[ 0,1\right] ^{n}\rightarrow R$
is convex. Then%
\begin{equation*}
E\left[ f\left( \mathbf{X}\right) \right] \leq E\left[ f\left( \mathbf{X}%
^{\prime }\right) \right] .
\end{equation*}%
If $f$ is permutation symmetric in its arguments and $\mathbf{\theta }\left(
k\right) $ denotes the vector $\mathbf{\theta }\left( k\right) =\left(
1,...,1,0,...,0\right) $ in $\left\{ 0,1\right\} ^{n}$, whose first $k$
coordinates are $1$ and whose remaining $n-k$ coordinates are zero, we also
have%
\begin{equation*}
E\left[ f\left( \mathbf{X}^{\prime }\right) \right] =\sum_{k=0}^{n}\binom{n}{%
k}\left( 1-\mu \right) ^{n-k}\mu ^{k}f\left( \mathbf{\theta }\left( k\right)
\right) .
\end{equation*}
\end{lemma}

\begin{proof}
A straightforward argument by induction shows that we can write any point $%
\mathbf{x}=\left( x_{1},...,x_{n}\right) \in \left[ 0,1\right] ^{n}$ as a
convex combination of the extremepoints $\mathbf{\eta }=\left( \eta
_{1},...,\eta _{n}\right) \in \left\{ 0,1\right\} ^{n}$ of $\left[ 0,1\right]
^{n}$ in the following way:%
\begin{equation*}
\mathbf{x}=\sum_{\mathbf{\eta }\in \left\{ 0,1\right\} ^{n}}\left(
\prod_{i:\eta _{i}=0}\left( 1-x_{i}\right) \prod_{i:\eta _{i}=1}x_{i}\right) 
\mathbf{\eta }.
\end{equation*}%
Convexity of $f$ therefore implies%
\begin{equation*}
f\left( \mathbf{x}\right) \leq \sum_{\mathbf{\eta }\in \left\{ 0,1\right\}
^{n}}\left( \prod_{i:\eta _{i}=0}\left( 1-x_{i}\right) \prod_{i:\eta
_{i}=1}x_{i}\right) f\left( \mathbf{\eta }\right) ,
\end{equation*}%
with equality if $\mathbf{x}\in \left\{ 0,1\right\} ^{n}$. Taking the
expectation and using independence and $E\left[ X_{i}\right] =\mu $ we get%
\begin{equation*}
E\left[ f\left( \mathbf{X}\right) \right] \leq \sum_{\mathbf{\eta }\in
\left\{ 0,1\right\} ^{n}}\left( \prod_{i:\eta _{i}=0}\left( 1-\mu \right)
\prod_{i:\eta _{i}=1}\mu \right) f\left( \mathbf{\eta }\right) .
\end{equation*}%
This becomes an equality if $\mathbf{X}$ is Bernoulli, for then $\mathbf{X}$
takes values only in $\left\{ 0,1\right\} ^{n}$. In particular $E\left[
f\left( \mathbf{X}\right) \right] \leq E\left[ f\left( \mathbf{X}^{\prime
}\right) \right] $, which gives the first assertion. If $f$ is permutation
symmetric then $f\left( \mathbf{\eta }\right) =f\left( \mathbf{\theta }%
\left( \left\vert \left\{ i:\eta _{i}=1\right\} \right\vert \right) \right) $
and we can rewrite the sum above as%
\begin{align*}
& \sum_{\mathbf{\eta }\in \left\{ 0,1\right\} ^{n}}\left( 1-\mu \right)
^{\left\vert \left\{ i:\eta _{i}=0\right\} \right\vert }\mu ^{\left\vert
\left\{ i:\eta _{i}=1\right\} \right\vert }f\left( \theta \left( \left\vert
\left\{ i:\eta _{i}=1\right\} \right\vert \right) \right) \\
& =\sum_{k=0}^{n}\binom{n}{k}\left( 1-\mu \right) ^{n-k}\mu ^{k}f\left(
\theta \left( k\right) \right) .
\end{align*}
\end{proof}

The next lemma is concerned with a series which can be viewed as a Rieman
sum approximating an instance of the Beta-function.

\begin{lemma}
\label{Beta Lemma}For $n\geq 2$ the sequence 
\begin{equation*}
c_{n}=\sum_{k=1}^{n-1}\frac{1}{\sqrt{k\left( n-k\right) }}
\end{equation*}%
satisfies $1\leq c_{n}\leq \pi $, and $c_{n}\rightarrow \pi $ as $%
n\rightarrow \infty $.
\end{lemma}

\begin{proof}
Define a function $\psi $ on $\left( 0,1\right) $ by 
\begin{equation*}
\psi \left( t\right) =\frac{1}{\sqrt{t\left( 1-t\right) }}.
\end{equation*}%
The change of variables $t\rightarrow \cos ^{2}\theta $ shows that 
\begin{equation*}
\int_{0}^{1}\psi \left( t\right) dt=\pi .
\end{equation*}%
It follows from elementary calculus that $\psi $ has a unique minimum at $%
t=1/2$ with minimal value $2$. This implies that $1/\sqrt{k\left( n-k\right) 
}\geq 2/n$ and therefore $c_{n}\geq 2\left( n-1\right) /n\geq 1$ for $n\geq
2 $. It also implies that the functions $\psi _{n}$ defined on $\left(
0,1\right) $ by%
\begin{equation*}
\psi _{n}\left( t\right) =\left\{ 
\begin{array}{ccccc}
\frac{1}{\sqrt{\frac{k}{n}\left( 1-\frac{k}{n}\right) }} & \text{if} & t\in 
\left[ \frac{k-1}{n},\frac{k}{n}\right) & \text{and} & k\leq n/2 \\ 
0 & \text{if} & t\in \left[ \frac{k-1}{n},\frac{k}{n}\right) & \text{and } & 
k-1\leq n/2<k \\ 
\frac{1}{\sqrt{\frac{k-1}{n}\left( 1-\frac{k-1}{n}\right) }} & \text{if} & 
t\in \left[ \frac{k-1}{n},\frac{k}{n}\right) & \text{and } & n/2<k-1%
\end{array}%
\right.
\end{equation*}%
satisfy $\psi _{n}\leq \psi $. Since 
\begin{equation*}
c_{n}=\sum_{k=1}^{n-1}\frac{1}{\sqrt{k\left( n-k\right) }}=\sum_{k=1}^{n-1}%
\frac{1}{n\sqrt{\frac{k}{n}\left( 1-\frac{k}{n}\right) }}=\int_{0}^{1}\psi
_{n}\left( t\right) dt
\end{equation*}%
this implies that $c_{n}\leq \pi $. Also $\psi _{n}\rightarrow \psi $ a.e.
so that by dominated convergence%
\begin{equation*}
c_{n}=\int_{0}^{1}\psi _{n}\left( t\right) dt\rightarrow \int_{0}^{1}\psi
\left( t\right) dt=\pi .
\end{equation*}
\end{proof}

\begin{proof}[Proof of Theorem \protect\ref{Main Theorem}]
If $X_{i}$ is trivial (i.e. if $\mu \in \left\{ 0,1\right\} $) (\ref{upper
bound}) is evident, so we can assume $\mu \in \left( 0,1\right) $. Define%
\begin{equation*}
f:\mathbf{x}\in \left[ 0,1\right] ^{n}\mapsto \exp \left( nKL\left( \frac{1}{%
n}\sum_{i=1}^{n}x_{i},\mu \right) \right) .
\end{equation*}%
Since the average is linear and $KL$ is convex (see \cite{CoverThomas}) and
the exponential function is nondecreasing and convex, the function $f$ is
also convex. $f$ is clearly permutation symmetric in its arguments. Lemma %
\ref{Convexity lemma} immediately gives 
\begin{equation}
E\left[ e^{nKL\left( M\left( \mathbf{X}\right) ,\mu \right) }\right] \leq E%
\left[ e^{nKL\left( M\left( \mathbf{X}^{\prime }\right) ,\mu \right) }\right]
=\sum_{k=0}^{n}\binom{n}{k}\left( 1-\mu \right) ^{n-k}\mu ^{k}f\left( 
\mathbf{\theta }\left( k\right) \right) .  \label{AUX I}
\end{equation}%
establishing also the first inequality in (\ref{upper bound}). Using the
special form of the function $f$ we find 
\begin{equation*}
f\left( \mathbf{\theta }\left( k\right) \right) =\exp \left( nKL\left( \frac{%
k}{n},\mu \right) \right) =\left( \frac{n-k}{n\left( 1-\mu \right) }\right)
^{n-k}\left( \frac{k}{n\mu }\right) ^{k}.
\end{equation*}%
Substitution in (\ref{AUX I}) leads to cancellation of the dependence in $%
\mu $ (proving the last statement of the theorem) and gives%
\begin{eqnarray*}
E\left[ e^{nKL\left( M\left( \mathbf{X}^{\prime }\right) ,\mu \right) }%
\right] &=&\sum_{k=0}^{n}\binom{n}{k}\left( \frac{k}{n}\right) ^{k}\left( 
\frac{n-k}{n}\right) ^{n-k} \\
&=&\frac{n!}{n^{n}}\sum_{k=1}^{n-1}\frac{k^{k}}{k!}\frac{\left( n-k\right)
^{n-k}}{\left( n-k\right) !}+2.
\end{eqnarray*}%
Using Stirling's formula (\ref{Stirling II}) and Lemma \ref{Beta Lemma} on
the last expression we obtain%
\begin{align*}
& E\left[ e^{nKL\left( M\left( \mathbf{X}^{\prime }\right) ,\mu \right) }%
\right] \\
& \leq \sqrt{2\pi n}\left( \frac{1}{e}\right) ^{n}e^{\frac{1}{12n}%
}\sum_{k=1}^{n-1}\frac{1}{\sqrt{2\pi k}\left( \frac{1}{e}\right) ^{k}}\frac{1%
}{\sqrt{2\pi \left( n-k\right) }\left( \frac{1}{e}\right) ^{n-k}}+2 \\
& =e^{\frac{1}{12n}}\sqrt{\frac{n}{2\pi }}\sum_{k=1}^{n-1}\frac{1}{\sqrt{%
k\left( n-k\right) }}+2 \\
& \leq e^{\frac{1}{12n}}\sqrt{\frac{\pi n}{2}}+2,
\end{align*}%
which gives (\ref{upper bound}). Similarly 
\begin{eqnarray*}
E\left[ e^{nKL\left( M\left( \mathbf{X}^{\prime }\right) ,\mu \right) }%
\right] &\geq &e^{-\frac{1}{6}}\sqrt{\frac{n}{2\pi }}\sum_{k=0}^{n}\frac{1}{%
\sqrt{k\left( n-k\right) }}+2 \\
&=&e^{-\frac{1}{6}}\sqrt{\frac{n}{2\pi }}c_{n}+2,
\end{eqnarray*}%
which gives (\ref{lower bound}) for Bernoulli variables.
\end{proof}

\section{Application to the PAC-Bayesian Theorem\label{Section PAC Bayes}}

Consider an unknown probability distribution $D$ on a set $\mathcal{Z}$, and
a set $\mathcal{H}$ of hypotheses $h:\mathcal{Z}\rightarrow \left[ 0,1\right]
$ (includes the loss function). To avoid a discussion of measurability $Z$
and $H$ are both assumed to be finite: Their cardinality is otherwise
irrelevant and will not appear in our results. The sample $\mathbf{S}=\left(
Z_{1},...,Z_{n}\right) $ is a $\mathcal{Z}^{n}$-valued random vector drawn
from $\Pr =D^{n}$. For $h\in \mathcal{H}$ we use $h\left( \mathbf{S}\right) $
to denote the $\left[ 0,1\right] $-valued random vector $h\left( \mathbf{S}%
\right) =\left( h\left( Z_{1}\right) ,...,h\left( Z_{n}\right) \right) $. We
write%
\begin{equation*}
h\left( D\right) =E_{z\thicksim D}\left[ h\left( z\right) \right] \text{ and 
}M\left( h\left( \mathbf{S}\right) \right) =\frac{1}{m}\sum_{i=1}^{m}h\left(
Z_{i}\right) .
\end{equation*}%
If $Q$ is a probability measure on $\mathcal{H}$, we write%
\begin{equation*}
Q\left( D\right) =E_{h\thicksim Q}\left[ h\left( D\right) \right] \text{ and 
}Q\left( \mathbf{S}\right) =E_{h\thicksim Q}\left[ M\left( h\left( \mathbf{S}%
\right) \right) \right] .
\end{equation*}%
The relative entropy of two probability measures $Q$ and $P$ on a set $%
\mathcal{H}$, denoted $KL\left( Q,P\right) ,$ is defined to be infinite if $%
Q $ is not absolutely continuous w.r.t. $P$. Otherwise, if $\frac{dQ}{dP}$
is the density of $Q$ w.r.t. $P$, we set 
\begin{equation*}
KL\left( Q,P\right) =E_{Q}\left[ \ln \frac{dQ}{dP}\right] .
\end{equation*}

\begin{theorem}
We have for any probability distribution $P$ on $\mathcal{H}$, for $n\geq 8$
and $\forall \delta >0$%
\begin{equation}
\Pr_{S}\left\{ \exists Q:KL\left( Q(\mathbf{S}),Q(D)\right) >\frac{KL\left(
Q,P\right) +\ln \frac{1}{\delta }+\ln \left( 2\sqrt{n}\right) }{n}\right\}
\leq \delta .  \label{PAC Bayes II}
\end{equation}
\end{theorem}

\begin{proof}
For every hypothesis $h\in \mathcal{H}$, applying the bound (\ref{simple
upper bound}) to the random vector $h\left( \mathbf{S}\right) $ gives%
\begin{equation*}
E_{\mathbf{S}}\left[ e^{nKL\left( M\left( h(\mathbf{S})\right) ,h(D)\right) }%
\right] \leq 2\sqrt{n}.
\end{equation*}%
Let $\mathbf{S}\mapsto Q_{\mathbf{S}}$ be any map from samples to the
probability distributions on $\mathcal{H}$ (a learning algorithm for Gibbs
classifiers). Using Jensen's inequality, convexity of the KL-divergence and
of the exponential function, we have%
\begin{align*}
& E_{\mathbf{S}}\left[ \exp \left( nKL\left( Q_{\mathbf{S}}(\mathbf{S}),Q_{%
\mathbf{S}}(D)\right) -KL\left( Q_{\mathbf{S}},P\right) \right) \right] \\
& \leq E_{\mathbf{S}}\left[ \exp \left( E_{h\thicksim Q_{\mathbf{S}}}\left[
nKL\left( M\left( h(\mathbf{S})\right) ,h(D)\right) -\ln \frac{dQ_{\mathbf{S}%
}}{dP}\left( h\right) \right] \right) \right] \\
& \leq E_{\mathbf{S}}\left[ E_{h\thicksim Q_{\mathbf{S}}}\left[ \exp \left(
nKL\left( M\left( h(\mathbf{S})\right) ,h(D)\right) -\ln \frac{dQ_{\mathbf{S}%
}}{dP}\left( h\right) \right) \right] \right] \\
& =E_{\mathbf{S}}\left[ E_{h\thicksim P}\left[ e^{nKL\left( M\left( h(%
\mathbf{S})\right) ,h(D)\right) }\left( \frac{dQ_{\mathbf{S}}}{dP}\right)
^{-1}\left( \frac{dQ_{\mathbf{S}}}{dP}\right) \right] \right] \\
& \leq E_{h\thicksim P}\left[ E_{\mathbf{S}}\left[ e^{nKL\left( M\left( h(%
\mathbf{S})\right) ,h(D)\right) }\right] \right] \\
& \leq 2\sqrt{n}.
\end{align*}%
Finally, by Markov's inequality,%
\begin{eqnarray*}
\delta &\geq &\Pr_{\mathbf{S}}\left\{ e^{nKL\left( Q_{\mathbf{S}}(\mathbf{S}%
),Q_{\mathbf{S}}(D)\right) -KL\left( Q_{\mathbf{S}},P\right) }>\frac{2\sqrt{n%
}}{\delta }\right\} \\
&=&\Pr_{\mathbf{S}}\left\{ KL\left( Q_{\mathbf{S}}(\mathbf{S}),Q_{\mathbf{S}%
}(D)\right) >\frac{KL\left( Q_{\mathbf{S}},P\right) +\ln \frac{1}{\delta }%
+\ln \left( 2\sqrt{n}\right) }{n}\right\} .
\end{eqnarray*}%
An appropriate worst-case choice of the function $S\mapsto Q_{\mathbf{S}}$
gives (\ref{PAC Bayes II}).\bigskip
\end{proof}

The loosest step in the proof is the use of Markov's inequality. The lower
bound (\ref{simple lower bound}) can be used to show that the other
inequalities are rather tight: Let $\mathcal{Z}$ be any large finite set, $%
\mathcal{H}$ a set of functions $h:Z\rightarrow \left\{ 0,1\right\} $ and $D$
a distribution such that the members of $\mathcal{H}$ all induce nontrivial
Bernoulli variables i.e. $E_{z\thicksim D}\left[ h\right] \in \left(
0,1\right) ,\forall h\in \mathcal{H}$. Let $P$ be uniform on $\mathcal{H}$.
For a sample $\mathbf{S}$ we define the posterior $Q_{\mathbf{S}}$ by its
density w.r.t. P:%
\begin{equation*}
\frac{dQ_{\mathbf{S}}}{dP}\left( h\right) =\frac{e^{mK\left( M\left( h\left( 
\mathbf{S}\right) \right) ,h\left( D\right) \right) }}{E_{h\thicksim P}\left[
e^{mK\left( M\left( h\left( \mathbf{S}\right) \right) ,h\left( D\right)
\right) }\right] }.
\end{equation*}%
Then%
\begin{equation*}
\psi \left( h,\mathbf{S}\right) =mK\left( M\left( h\left( \mathbf{S}\right)
\right) ,h\left( D\right) \right) -\ln \frac{dQ_{\mathbf{S}}}{dP}=\ln
E_{h\thicksim P}\left[ e^{mK\left( M\left( h\left( \mathbf{S}\right) \right)
,h\left( D\right) \right) }\right]
\end{equation*}%
is independent of $h$. Therefore with (\ref{simple lower bound}) 
\begin{eqnarray*}
E_{\mathbf{S}}\left[ e^{E_{h\thicksim Q}\left[ mK\left( M\left( h\left( 
\mathbf{S}\right) \right) ,h\left( D\right) \right) -\ln \frac{dQ_{S}}{dP}%
\right] }\right] &=&E_{\mathbf{S}}\left[ e^{E_{h\thicksim Q}\left[ \psi
\left( h,\mathbf{S}\right) \right] }\right] \\
&=&E_{\mathbf{S}}\left[ E_{h\thicksim Q}\left[ e^{\psi \left( h,\mathbf{S}%
\right) }\right] \right] \\
&=&E_{h\thicksim P}\left[ E_{\mathbf{S}}\left[ e^{mK\left( M\left( h\left(
S\right) \right) ,h\left( D\right) \right) }\right] \right] \\
&\geq &\sqrt{n}.
\end{eqnarray*}%
This can be rewritten as the statement: For every $\delta >0$ 
\begin{equation*}
E_{\mathbf{S}}\left[ \exp \left( mE_{h\thicksim Q}\left[ K\left( M\left(
h\left( \mathbf{S}\right) \right) ,h\left( D\right) \right) \right]
-KL\left( Q,P\right) +\ln \frac{1}{\delta }+\ln \sqrt{m}\right) \right] \geq
\delta ,
\end{equation*}%
a very weak lower bound version of the PAC-Bayesian theorem, which
nevertheless shows that an elimination or the $\sqrt{m}$ term, if it is at
all possible, would have to follow a completely different path.

\end{document}